\documentclass[sigconf]{acmart}

\AtBeginDocument{%
  }

\setcopyright{acmlicensed}
\copyrightyear{2025}
\acmYear{2025}
\acmDOI{XXXXXXX.XXXXXXX}

\acmConference[KDD '25]{31st SIGKDD Conference on Knowledge Discovery and Data Mining (KDD ’25),
August 03–07, 2025, Toronto, Canada.}{August 03–07,
  2025}{Toronto, Canada}

\usepackage{libertine}
\usepackage{listings}
\usepackage{enumitem}
\usepackage{alltt}
\usepackage{multirow}

\definecolor{delim}{RGB}{0,0,0}
\definecolor{numb}{RGB}{0,0,0}
\definecolor{string}{RGB}{0,0,0}

\lstdefinelanguage{json}{
    showspaces=false,
    showtabs=false,
    breaklines=true,
    showstringspaces=false,
    breakatwhitespace=true,
    basicstyle=\ttfamily\small,
    upquote=true,
    morestring=[b]",
    stringstyle=\color{string},
    literate=
     *{0}{{{\color{numb}0}}}{1}
      {1}{{{\color{numb}1}}}{1}
      {2}{{{\color{numb}2}}}{1}
      {3}{{{\color{numb}3}}}{1}
      {4}{{{\color{numb}4}}}{1}
      {5}{{{\color{numb}5}}}{1}
      {6}{{{\color{numb}6}}}{1}
      {7}{{{\color{numb}7}}}{1}
      {8}{{{\color{numb}8}}}{1}
      {9}{{{\color{numb}9}}}{1}
      {\{}{{{\color{delim}{\{}}}}{1}
      {\}}{{{\color{delim}{\}}}}}{1}
      {[}{{{\color{delim}{[}}}}{1}
      {]}{{{\color{delim}{]}}}}{1},
}

\begin{document}

\title{Beyond the Surface: Uncovering Implicit Locations with LLMs for Personalized Local News}

\author{Gali Katz}

\affiliation{%
  \institution{Taboola}
  \city{Tel Aviv}
  \country{Israel}}
\email{gali.k@taboola.com}

\author{Hai Sitton}
\affiliation{%
  \institution{Taboola}
  \city{Tel Aviv}
  \country{Israel}}
\email{hai.s@taboola.com}

\author{Guy Gonen}
\affiliation{%
  \institution{Taboola}
  \city{Tel Aviv}
  \country{Israel}}
\email{guy.gonen@taboola.com}

\author{Yohay Kaplan}
\affiliation{%
  \institution{Taboola}
  \city{Tel Aviv}
  \country{Israel}}
\email{yohay.k@taboola.com}

\renewcommand{\shortauthors}{Katz et al.}

\begin{abstract}

News recommendation systems aim to serve users with relevant content. One key application is homepage personalization, where articles shown in the news homepages are tailored to individual users. While homepage personalization has potential for higher engagement, there are qualitative aspects impacting recommendation such as content type, editorial stance, and geographic focus. Local newspapers, in particular, aim to engage readers from specific cities or regions by balancing local, national, and international coverage. Identifying local articles presents a challenge, as implicit locations are often conveyed through references to local slang, notable figures, sports teams, or landmarks rather than being explicitly mentioned.

Traditional approaches, such as Named Entity Recognition (NER), extract entities from text, while Knowledge Graphs help infer locations that may not be explicitly stated. The emergence of Large Language Models (LLMs) has introduced new possibilities, raising questions about their ability to internalize relational knowledge and potentially replace traditional methods. However, these advancements also raise concerns regarding hallucinations, outdated data and a lack of explainability.

This paper presents a novel approach to classifying local articles using LLMs within Taboola's "Homepage For You" personalization algorithm. We compared traditional classification techniques with LLM-based models, evaluating both standalone LLMs and those augmented with Knowledge Graph data. Our analysis revealed three key insights: (1) Knowledge Graphs enhance NER models' ability to detect implicit locations, (2) LLMs outperform traditional methods, and (3) LLMs can effectively identify local content without requiring Knowledge Graph integration.

Offline evaluations showed that LLMs excel in implicit local classification tasks compared to traditional methods, while online A/B tests showed an increase in user engagement metrics.

Finally, we built and deployed a scalable pipeline that integrates location classification into our production news recommendation system.
At full scale, this deployment resulted in an average uplift of 27\% in local articles served, maintaining newspapers' brand identity.
By leveraging state-of-the-art auxiliary models, our framework enables real-time indexing of article location data, making it readily available for training and significantly enhancing homepage personalization.

\end{abstract}




\received{03 February 2025}

\maketitle

\section{Introduction}

News websites are a popular medium for consuming the latest headlines and updates. One of their main challenges is maximizing reader engagement with content. 
News have to be \textbf{fresh}, and to present what is currently \textbf{popular} or \textbf{need-to-know} by the user. 

Traditional websites rely on editors to manually curate and position content, resulting in a static experience that is identical for all users. 
Today, recommender systems can serve different content to each user while adhering to editorial guidelines. This has led to a growing trend of personalized homepages that deliver tailored content to every user.

Homepage personalization enables news websites to present content which is correlated with each reader's interests, enhancing their experience and increasing overall engagement \cite{netflix, semerci2019homepage}. Most homepages typically follow a consistent structure for easy navigation. The header contains the site logo, brand name, and navigation menu. Below this is a \textit{hero} section featuring the latest or most important news, displaying a title, thumbnail, and brief description without topic constraints. The site then presents themed sections such as sports, politics, news, and opinions. These sections each focus on a specific topic. News websites vary in their geographic scope, serving local, national, or international audiences.

While prediction accuracy is the most discussed property in recommendation system literature \cite{ricci2021recommender}, Homepage Personalization requires us to consider additional performance measures for our recommender system, including recency, diversity, location-aware recommendations, and adaptivity to user preferences.

Local news websites serve readers in specific geographic areas with news, events, and stories that directly impact their communities. Through strong connections with local businesses and organizations, these sites provide comprehensive coverage of neighborhood happenings. Our current work will focus on serving personalized local news in digital newspapers that support homepage personalization.

Personalized homepages of local websites face the challenge of accurately identifying and promoting relevant content to users. While \textbf{\textit{relevance}} takes several forms, such as \textit{popularity} versus \textit{personalization}, personalizing local news websites faces a specific challenge: promoting local content rather than simply optimizing for standard KPIs like click-through rate or watch time. This challenge stems from data sparsity—local news typically has smaller audiences than national or international news, (local U.S. newspapers are declining, due to a decrease in readership and newsroom staff \cite{hayes2018decline, martin2019local}), and the abundance of non-local news articles allows to cherry pick the most engaging articles which are not limited by geography. Consequently, the model tends to show bias toward non-local content. In this work, we address this challenge by incorporating both article and user locality information into our recommendation system model.

When addressing the extraction of entities from text, particularly geographic locations, traditional Named Entity Recognition (NER, \cite{grishman1996message, sang2003introduction}) models were until recently a standard choice in recommendation systems \cite{de2015semantics, fudholi2023bert, li2020deep}. However, NERs lack a built-in \textit{entity linking} mechanism—also known as \textit{named entity disambiguation} \cite{de2015semantics}—making it difficult to determine the precise identity of entities mentioned in text. Thus, NERs might present poor performance when encountering:

\begin{enumerate}
\item Implicit locations—such as in sports team names (e.g., "The Dolphins," which is Miami's football team).
\item Geographical ambiguity—or toponyms (place names)—
occurs when a city name exists in multiple states (e.g., Hollywood, Alabama or Hollywood, California). Disambiguating place names is a well-known challenge in Geographic Information Retrieval (GIR), and is also recognized in news articles and gazetteers \cite{garbin2005disambiguating}. Common disambiguation approaches include map-based, knowledge-based, or machine learning based methods \cite{buscaldi2011approaches}. 
\end{enumerate}

Knowledge Graphs (KGs) \cite{ehrlinger2016towards} solve the entity linking limitation of NERs by helping detect implicit locations and have widespread applications in recommender systems \cite{guo2020survey}. These graphs organize information in triplets: (head entity, relation, tail entity). Entities—such as persons, locations, or organizations—are represented as nodes, while their relationships (e.g., "located in," "born in," "is a member of") form edges. For example, the triplet (Skylar Thompson, member of, Miami Dolphins) shows that Skylar Thompson (person entity) is a member of the Miami Dolphins (organization entity), and the Miami Dolphins are based in Miami (location entity). Due to this connection, a sports article about Skylar Thompson can be categorized as local news on a Miami website.

Built on extensive datasets, Knowledge Graphs provide current, reliable information and domain-specific knowledge. They also make recommendation results more explainable \cite{wang2018ripplenet}. However, updating a Knowledge Graph can be costly and often requires manual work. 

As LLMs emerged, many recommender systems sought to leverage their general knowledge and ability to generalize for solving complex challenges \cite{zhao2023recommender, lin2023can}. However, these systems face several limitations: they can become outdated due to expensive training costs, lack interpretability (especially when not applying the Chain-of-thought \cite{wang2022self} technique), show bias, and produce hallucinations \cite{pan2024unifying}.

This paper compares different methods for classifying local articles and examines the added value each method offers for detecting implicit locations in text. We then deploy the preferred option in an online recommender system. 

\section{Related Work}
\textbf{\textit{Traditional methods.}} Until recently, news recommendation systems treated location-based articles primarily as an entity extraction problem. These systems identified locations both explicitly through Named Entity Recognition \cite{de2015semantics, fudholi2023bert, li2020deep} and implicitly through Knowledge Graphs \cite{wang2018ripplenet, wang2019multi, wang2018dkn, wang2019exploring}.

Several studies have tackled the specific challenge of implicit locations. \cite{son2013location} noted that news articles frequently use vague location references, especially when mentioning only cities or suburbs. The authors developed Explicit Localized Semantic Analysis (ELSA), which uses geo-tagged documents to characterize locations and maps both these documents and news articles into a topic space using Explicit Semantic Analysis (ESA) \cite{gabrilovich2009wikipedia}. The system then analyzes Wikipedia-based topic vectors to determine the relevance between a user's location and news articles, recommending the most relevant content.

\textbf{\textit{BERT-based methods.}} Research then explored whether BERT \cite{kenton2019bert} model possess an \textit{inherent relational knowledge}, thus might be able to detect location signals. \cite{petroni2019language} demonstrated that BERT achieves relational understanding comparable to traditional NLP systems with oracle-level knowledge, even without fine-tuning. Moreover, BERT exceled at open-domain question answering compared to supervised baselines (e.g., "Dante was born in [Mask]", \cite{petroni2019language}).

UNBERT \cite{zhang2021unbert} advanced this concept by explicitly building a model to process word-level signals such as locations. They introduced a pre-trained BERT model for news recommendation designed to address the cold-start problem using \textit{location-based signals} at the word level. Their model represented users through their browsed news text and employed two modules: a \textit{Word-Level} Module (WLM) with transformer layers to calculate hidden representations and propagate word-level matching signals, and a \textit{News-Level} Module to capture matching at the news level. While this was the first model to successfully capture news location signals, it required locations to be \textbf{explicitly} mentioned in the article's text.

\textbf{\textit{LLMs enrichment methods.}} Recent advancements in LLMs like ChatGPT \cite{open2023chatgpt, qin2023chatgpt} present a promising alternative to Named Entity Recognition models and Knowledge Graphs. These models demonstrate strong capabilities in contextual understanding and reasoning. Unlike conventional approaches, they are not restricted by fixed entity-relation schemas, allowing for flexible and dynamic relational modeling. 

Several studies have explored ways to incorporate Knowledge Graphs to enhance LLMs' knowledge. \cite{zhang2019ernie} developed a model that uses Knowledge Graph structures through knowledge embeddings \cite{bordes2013translating} as input to a BERT-based model. To better integrate textual and Knowledge Graph information, they created a new objective: the model predicts masked named entity alignments within text, learning to match entities with their correct graph context. This results in a more knowledge-enriched representation.
    
Further, \cite{liu2020k} enhanced BERT by integrating KG triplets into sentences but found that this approach disrupted readability and structure. To address this, they introduced a soft-positioning technique and a masked transformer layer that preserved sentence integrity while masking KG information. They also introduced the term Knowledge Noise (KN)—irrelevant or misleading KG data—and showed that the suggested soft-positioning and masking techniques helped reduce it. Lastly, they suggested refining KG triplet selection to further minimize noise.

\cite{liu2021kg} proposed a KG-augmented decoder that incorporates hierarchical graph structures during decoding. According to the authors, this approach enables the model to better capture relationships between concepts and their neighbors, improving output accuracy. In their work, graph attention mechanisms were employed to aggregate rich semantic information, enhancing the model’s capacity to generalize across unfamiliar concepts.
   
Other research examined methods for aligning KG information with LLM predictions. \cite{yasunaga2022deep} used (text, local KG) pairs for LLM pre-training, and \cite{zhang2022greaselm} combined Pre-trained Language Models (PLMs) and Graph Neural Network (GNN) representations via layered modality interactions. \cite{feng2023knowledge} introduced the Knowledge Solver (KSL), which guides LLMs in structured KG searches by transforming retrieval tasks into multi-hop decision sequences, empowering zero-shot knowledge retrieval.
    
To this date, no existing work has, to our knowledge, specifically applied LLMs' ability to reason about implicit locations within a news recommendation system.

\section{Method}
We evaluated our local articles classification using four methods: Standalone-NER, NER with Knowledge Graph, Standalone-LLM, and a combined approach that integrates KG information into the LLM prompt. For our NER implementation, we employed a standard BERT-based model to identify location entities in the text. We used ChatGPT \cite{open2023chatgpt} (specifically the gpt-3.5-turbo-0301 snapshot) as our LLM, and Wikidata \cite{vrandevcic2014wikidata} — a free, open knowledge base with 115M data items maintained by both human and machine editors — as our Knowledge Graph.

\subsection{Dataset}
We collected 33,880 unique news headlines and article descriptions (the first paragraph of each article) from 142 English-language local newspapers during the week of September 13–20, 2024.

The dataset was processed through: Standalone-NER, 
KG-Enriched NER (i.e. with Knowledge Graph information), 
Standalone-ChatGPT and KG-Enriched ChatGPT (see prompt in Appendix \ref{appendix:prompt}).

The items sent to KG-Enriched ChatGPT consisted of Standalone-ChatGPT's False Positives (415) and False Negatives (498), totaling 913 items that were enriched with Knowledge Graph triplets (see confusion matrices of ChatGPT and KG-Enriched-ChatGPT in Appendix \ref{appendix:confusion_matrix}).

\subsection{NER Enriched by Knowledge Graph}
Initially, NER identified explicit entities in the article texts. 
These entities included persons, locations, organizations, and miscellaneous items (e.g., Person-Jyilek, Person-Myles, Organization-West Virginia State University, Location-Georgia). The findings were then sent to the Wikidata Knowledge Graph to detect implicit locations in the text. Each NER tag was processed differently:

\begin{itemize}

\item \textbf{Location and Organization entities}:
Wikidata organizes information into items, with each item described using predefined properties. Each property is mapped to a value, creating a structured and organized dataset. This structure allows us to build a knowledge graph where items are nodes and properties become edges. In our case, we utilized the P131 relation property, which denotes an entity's administrative territorial location. This property encompasses geographical entities such as cities, organizations, landmarks, and event locations. Location data was retrieved and we traversed the graph up to 4 hops, following the P131 relation structure: neighborhood → city → county → state → country. 

\begin{itemize}
\item \textbf{\textit{Classification strategy}}: An article was classified as local, if its associated entity matched the newspaper’s city or state using the P131 property. Otherwise, we iteratively traversed the P131 hierarchy until we reached an entity whose P131 matches the newspaper's city or state.
\end{itemize}

\item \textbf{Person entity:} We validated person entities by confirming that the "instance of" property (P31: human) correctly identified each entity as a person in NER. We then extracted the entity description and sports team affiliation property (P54). While we evaluated other geographical properties—including "place of birth" (P19) and "residence" (P551)—we ultimately excluded them due to their negative impact on coverage. This aligns with \cite{liu2020k}'s definition of Knowledge Noise. For example, an article about Taylor Swift is unlikely to be contextually relevant to her birthplace or residence. Thus, we don't consider these locations when classifying an article as local.

\begin{itemize}
\item \textit{\textbf{Classification Strategy}}: An article was classified as local if the newspaper's city or state appeared in the entity’s description or sports team affiliation (for athlete entities).
\end{itemize}
\end{itemize}

\subsubsection{Entity Linking}\label{entity_linking}
We chose a naive approach to link between the article’s location entity and the location of the website:

\begin{itemize}
\item If the entity has a population relation (P1082), select the entity with the highest population.
\item Otherwise, choose the entity with the most relations (probably the most popular one).
\end{itemize}

\subsection{Standalone-ChatGPT}
\subsubsection{Prompt Engineering}
We instructed ChatGPT to act as an editor and identify city, state, and country entities using the In-Context Learning technique \cite{brown2020language}. Following the Chain-of-Thought (CoT) paradigm \cite{wang2022self}, we created a step-by-step guide to resolve implicit locations. We then requested the output in valid JSON format (noting that ChatGPT 3.5 lacks JSON schema validation) and provided examples at the end (i.e., few shot \cite{brown2020language}). 
To handle ChatGPT's tendency to generate multiple location outputs for a single article, we refined the prompt (see Appendix \ref{appendix:prompt}) to select only the first detected location entity.

\subsection{KG-Enriched ChatGPT}
\subsubsection{Prompt Enrichment with Knowledge Graph Data}
We added the following instruction to the prompt: 
\begin{lstlisting}[language=json,firstnumber=1]
When a Knowledge Graph provides a triplet (entity, relation, entity), use it to determine the location related to the article.
\end{lstlisting}

For example, for the article with the title and description below: 
\begin{lstlisting}[language=json,firstnumber=1]
{
  "title": "Republican operative admits paying Artiles for opo research on trial's first day",
  "description": "GOP consultant Patrick Bainter testified that he paid Frank Artiles $15,000 a month and sent $100,000 to a political action committee for background information on incumbent Jose Javier Rodriguez."
}
\end{lstlisting}
NER's and KG's outputs were: 
\begin{lstlisting}[language=json,firstnumber=1]
{
    "NER": "Person-Frank Artiles,Person-Patrick Bainter,Miscellaneous-Republican,Person-Jose Javier Rodriguez"
    "KG": "[('Frank Artiles', 'personDescription', 'florida state representative'),('Jose Javier Rodriguez','personDescription','Puerto Rican politician')]"
}
\end{lstlisting}

Thus, the Knowledge Graph found only two persons out of three mentioned in the text: \textit{Frank Artiles} and \textit{Jose Javier Rodriguez}.

The input to the LLM included the \textbf{title}, \textbf{description}, and \textbf{triplet} of the KG. 
In this specific example, the relation label personDescription: \textit{florida state representative} seemed to helped retrieving the correct locations -- without this knowledge, the LLM returned null values.

Below is the LLMs response to this specific article:
\begin{lstlisting}[language=json,firstnumber=1]
{
  "title": "Republican operative admits paying Artiles for opo research on trial's first day",
  "item_city": "miami",
  "item_state": "florida",
  "item_country": "united states"
}
\end{lstlisting}

Moreover, the location of this article was implicit (i.e., not explicitly mentioned) and was equal to the newspaper's location which is Miami, Florida, thus it was classified as local.

\subsubsection{ChatGPT Validations and Consistency}\label{chatGPT_validations}
We validated that each city returned by ChatGPT was located within its stated state and that each state was within its stated country. For consistency, we ran the prompts through ChatGPT three separate times and merged the results. The default temperature was used to maintain ChatGPT's natural output in this case.

\section{Offline Analysis}
\subsection{Evaluation}
We evaluated the models using F1-Score, precision, and recall metrics. Our goal was to display local articles in designated slots on local newspaper homepages. Since these slots require only a few articles, we could maintain sufficient local article visibility by identifying a small number of articles but to do so with high precision.

\subsubsection{Ground Truth Validation}
Our ground truth was based on manual tagging performed by editors. Editors use tags to optimize search engines and enable user searches, giving them a clear incentive to maintain tag accuracy. Articles were tagged as \textbf{\textit{local}} or \textbf{\textit{national}}, with untagged articles excluded from the analysis. In our post-processing of the LLM results, we classified articles as local when ChatGPT's returned city/state matched the newspaper's city/state, and national otherwise.

\subsubsection{Detecting Local Newspaper Geographic Affiliations}\label{newspaper_geographic}
Taboola does not store information about newspaper locations. To determine a newspaper's city and state, we identified each newspaper's location by analyzing where the majority of its website visitors were based.

\subsection{Models Performance}\label{models_performance}

Table 1 presents the model's performance when classifying local articles. The KG-Enriched ChatGPT outperformed all other models with the best F1-Score of 81.82\% and precision of 82.12\% and detected 2.93\% more local items than Standalone-ChatGPT.

Standalone-NER achieved relatively high precision (74.55\%) but suffered from poor recall (41.01\%), as it relies solely on explicit locations. However, augmenting NER with Knowledge Graph enrichment improved its recall by 22\% to 63.04\%. The improved recall likely stems from the KG’s ability to identify implicit locations in local articles (see Figure 1).

\begin{table}
  \caption{Models performance for local items classification task}
  \label{tab:perf}
  \begin{tabular}{cccl}
    \toprule
    Model&F1 Score&Precision&Recall\\
    \midrule
    Standalone-NER & 52.94\% & 74.55\% & 41.04\% \\
    KG-Enriched NER& 68.31\% & 74.54\% & 63.04\%\\
    Standalone-ChatGPT& \underline{80.02}\% & \underline{81.51}\% & \underline{78.59}\%\\
    KG-Enriched ChatGPT& \textbf{81.82}\% & \textbf{82.12}\% & \textbf{81.52}\%\\
  \bottomrule
\end{tabular}
\end{table}

\subsection{Implicit Locations Recall}
The editors of local newspapers tagged 2,327 unique articles as local, which represent 6.87\% of the entire dataset. We then determined whether each article was considered implicit as follows: 

An article was considered \textbf{implicit} if the newspaper tagged it as local but the NER model did not identify it as such. NER classified an article as local if its location entities (city/state) matched the newspaper's city/state, as described in section \ref{newspaper_geographic}

\begin{figure}[ht]
  \centering
  \includegraphics[width=\linewidth]{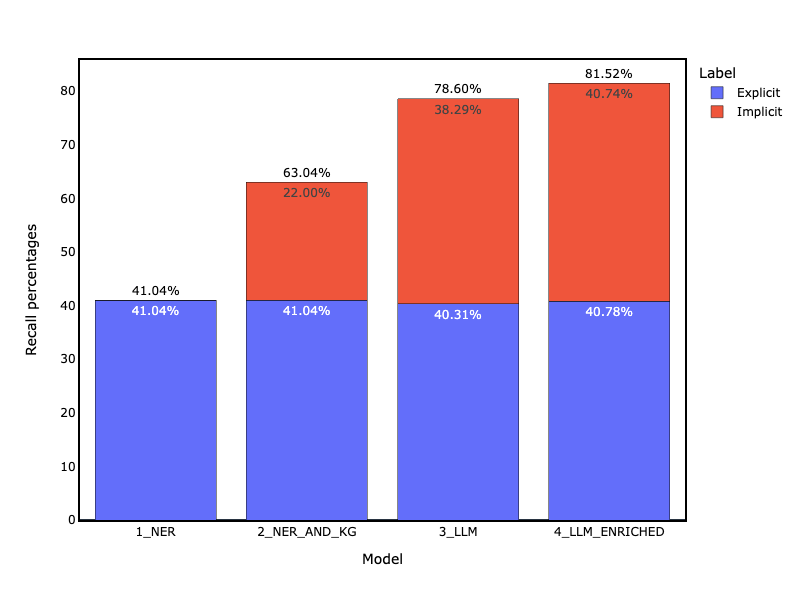}
  \caption{Local Articles Recall}
  \Description{Recall percentages of articles with local content, analyzed by four entity extraction models and categorized by explicit versus implicit location references}
  \label{fig:recall}
\end{figure}

Figure \ref{fig:recall} presents the recall of local articles. KG-Enriched NER identified 22\% more articles with implicit locations, beyond the 41.04\% explicit locations detected by Standalone-NER. ChatGPT discovered 38.29\% of implicit articles, while KG-Enriched ChatGPT improved the recall to 40.74\% implicit articles.

The KG-Enriched GPT improved implicit location identification by only 2.45\%, a marginal gain that calls into question its overhead implementation.

\subsection{What Caused ChatGPT to Misclassify?}
We found four main causes of incorrect classifications by ChatGPT (enriched and non enriched):
\begin{enumerate}
    \item Editor tagging human errors: Articles with local locations were sometimes tagged as national content by editors.
    \item County-state confusion: The JSON output format lacked a specific county field, causing ChatGPT to misidentify counties as states.
    \item Knowledge noise: In KG-enriched ChatGPT, irrelevant knowledge graph data interfered with accurate classification.
    \item Oversimplified criteria: Our approach of using either a city or a state match to determine local content led to some classification errors.
\end{enumerate}

\subsection{Knowledge Graph Enriched Prompt}
KG-enriched prompt increased the number of local items that were discovered by 3\%. 
We found evidence of Knowledge Noise \cite{liu2020k} and also examined whether Toponyms Ambiguity was present in our data and if enrichment caused the ambiguity or helped disambiguate any discovered toponyms \cite{garbin2005disambiguating}.

\subsubsection{Knowledge Noise}\label{kn}
Overall enrichment helped minimizing False Negatives and False Positives. However, there were cases of Knowledge Noise.
Here is a story from a newspaper from Raleigh, North Carolina:
\begin{lstlisting}[language=json,firstnumber=1]
{
    "title": "Javion Magee gave homeless person $228, had rope coiled around his neck, sheriff says",
    "description":"The 21-year-old's body was discovered in Henderson. His death raised suspicion of lynching, while investigators point to suicide."
    "triplet":"[(County, countyLabel, Clark County), (State, regionLabel, North Carolina), (State, stateLabel, Nevada), (Country, additionalRegionLabel, United States)]"
}
\end{lstlisting}
Standalone-ChatGPT returned the location Henderson, North Carolina, while KG-Enriched ChatGPT returned Henderson, Nevada. This problem stemmed from our Knowledge Graph disambiguation heuristic of selecting the place with the highest population—in this case Henderson, Nevada (330,000 people in 2023). Unfortunately, the crime occurred in the smaller city of Henderson, North Carolina (14,000 people in 2023).

\subsubsection{Toponym Ambiguity}
We marked articles as "ambiguous" when the model identified the same city in different states. Results showed that Toponym Ambiguity was rare, occurring in 3.10\% of Standalone-ChatGPT cases and 3.18\% of KG-Enriched ChatGPT cases. 
There were cases that enrichment created Toponyms (see Knowledge Noise example in \ref{kn}) and there was also evidence for \textbf{Toponym disambiguation}, like this one from a newspaper from Sarasota, Florida:
\begin{lstlisting}[language=json,firstnumber=1]
{
  "title": "Man receiving medical care, pending physical arrest after shooting incident near Charlotte/Sarasota County Border",
  "description":"A suspect is receiving medical care, pending physical arrest, after a shooting incident near the County line of Charlotte and Sarasota",
}
\end{lstlisting}

Standalone-ChatGPT returned that the location of the article is Charlotte, North Carolina, while KG-Enriched GPT returned Charlotte, Florida, based on the following triplet information:
\begin{lstlisting}[language=json,firstnumber=1]
[('County', 'countyLabel', 'Sarasota County'), ('State', 'stateLabel', 'Florida'), ('Country', 'additionalRegionLabel', 'United States')].
\end{lstlisting}

Since we classified an article as local when its city/state matched the newspaper's location, this article was classified as local to Florida.

\section{Deployment}
\begin{figure*}[ht]
  \centering
  \includegraphics[width=\linewidth]{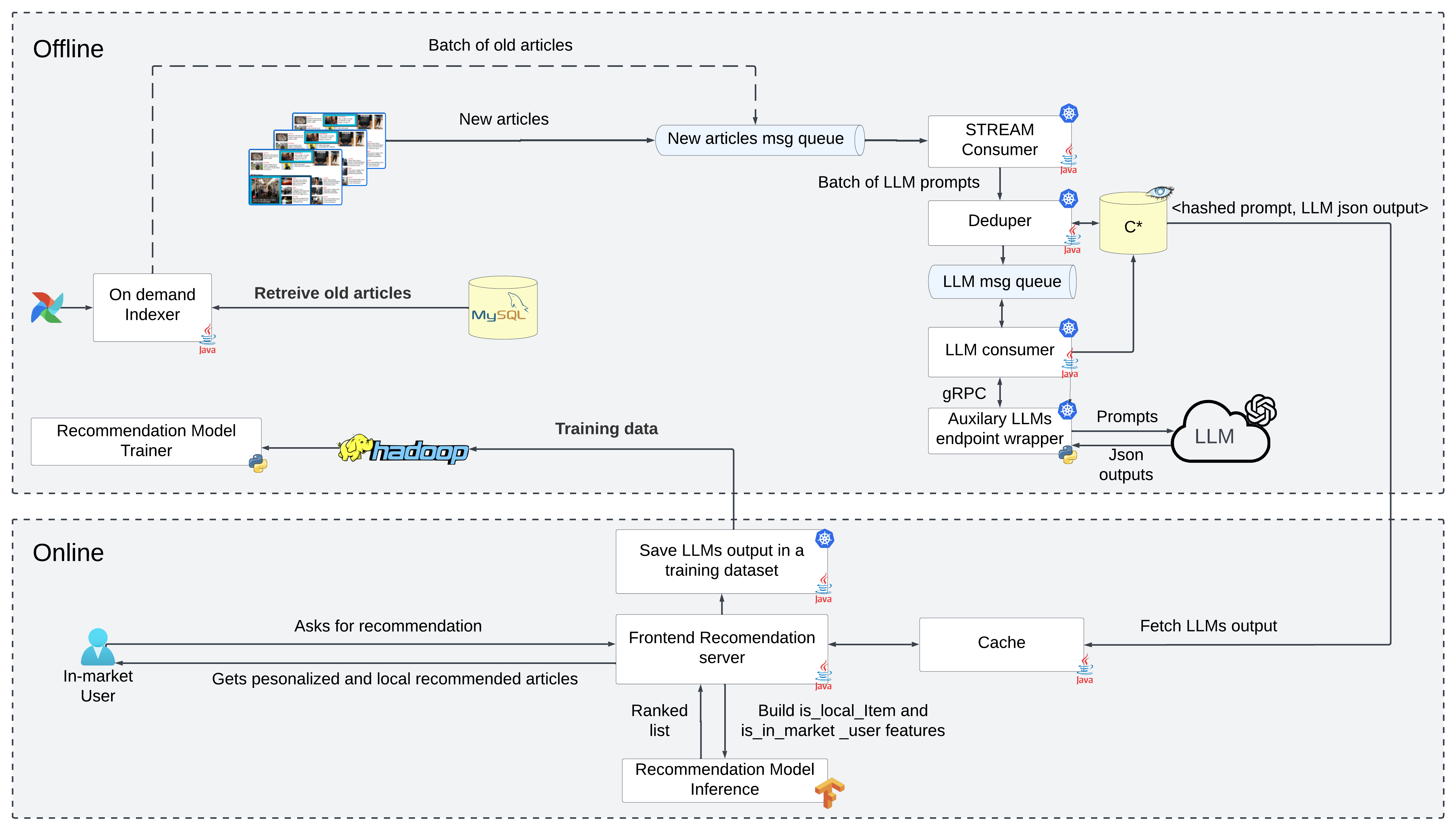}
  \caption{Localization with LLMs Production Pipeline}
  \label{fig:deployment}
  \Description{The production deployment architecture, divided into offline and online flows.}
\end{figure*}

The scalable pipeline we built at Taboola enables us to leverage auxiliary models—including state-of-the-art LLMs and our own in-house models. The pipeline is \textit{asynchronous} because inference times for different LLMs are typically slower than our data streaming pace. To handle this, we used multiple message queues throughout the pipeline. The architecture is split in two logical flows: \textbf{\textit{Offline}} and \textbf{\textit{Online}}, as shown in Figure \ref{fig:deployment}.

\subsection{Offline Flow} 
Our system listens through a message queue for newly crawled articles. It then extracts features for use by the LLM through \textbf{STREAM} (\textbf{S}ystem for \textbf{T}ailored \textbf{R}ecommendation and \textbf{A}uxiliary \textbf{M}odels) consumer service. 

In the local news use case, the STREAM consumer extracts the title, description, and editorial tags from each new article. It builds a batch of prompts suitable for the LLM and sends them to a Deduper service, to avoid sending duplicate requests. The Deduper stores entries in the key-value store Cassandra (C*), where the key is the hashed prompt and the value remains \textit{out-of-vocabulary} until a result returns from the LLM. 

The batch of prompts is then sent to another queue. The LLM Consumer receives the prompt batches and sends them to an Auxiliary LLMs Endpoint Wrapper service that calls the LLM's API. The result from the LLM (a JSON output in case of the location classification) is stored by the LLM Consumer back in C*.

For cases requiring processing older items, we support triggering an offline manual job via Airflow that fetches old items from databases such as MySQL and injects them back into the New Articles Message Queue. From there, they enter the indexing pipeline.

\subsection{Online Flow}
When a user requests a recommendation, the Recommendation Server fetches LLM outputs from a cache above the C* key-value store. The server then builds relevant features required for a personalized model, taking into account the \textbf{user features}, the \textbf{context features} and the \textbf{item features}.

Using TensorFlow Serving, an \textbf{inference} is performed on the recommendation model. A personalized ranked list optimized for \textbf{Click-Through-Rate} is generated, sorted from most to least relevant for this specific user within this specific context, including local considerations. 

\subsection{Training} The model trains on LLM outputs. During the online phase, we store the calculated feature values in HDFS storage. The Recommendation Model Trainer in offline mode can then fetch the dataset containing these values and train several times a day.

\section{Online Results}\label{sec:online_engagement_results}

\subsection{Modeling}
In order to embed the LLMs location-based knowledge we created two features. One is an article level feature, the other is a user feature.

\begin{enumerate}
   \item \verb|is_local_item| - derived from the LLM result and cross-referenced with the newspaper's location (as described in subsection \ref{newspaper_geographic}).
   \item \verb|is_in_market_user| - A user was considered in-market if:
   \begin{enumerate}
     \item He was based in the US and had a DMA code that matched the newspaper's location DMA code.
     \item He was not based in the US (no DMA code) but their geographic affinity matched the newspaper's location.
   \end{enumerate}
 \end{enumerate}

We distinguished between US and non-US users because DMA codes provided more accurate results in our offline analyses, thus we prioritized them when available.
Finally, we incorporated both article locality and user in-market-ness information into Taboola recommender system model \cite{zeldes2017deep}. This model than predicted the optimal CTR for a specific article and user, given the current context. 

\subsection{Engagement}
The results in this section were collected over 3 months of recommendations from local newspapers with "Homepage For You" logic enabled. The data was user sticky (i.e., each variant was assigned to a different group of users) and included clicks, views, and users. We conducted an A/B test with treatment and control groups. The control group did not use LLMs for local article classification, nor was this classification data integrated into the recommendations model.

\begin{table}
  \caption{Effects on In Market Users}
  \label{tab:engagement}
  \begin{tabular}{cccl}
    \toprule
    Metric&Mean Diff&Pooled std&t-Test\\
    \midrule
    Local Views Per User&14.95&32.83&significant\\
    Clicks Per User&0.06&0.51&non significant\\
  \bottomrule
\end{tabular}
\end{table}

Table \ref{tab:engagement} shows the user engagement results, examining local views per user and overall clicks per user for in-market users. A two-sample independent t-test was conducted to compare the number of views per user and clicks per user between the variant and control groups. Results indicated a significant difference in local views per user between the variant (M = 70.52, SD = 39.12) and the control group (M = 55.57, SD = 24.99), with the variant group having more views, \textit{t}(222) = 3.39, \textit{p} < .001, \textit{d} = 0.46. However, there was no significant difference in overall clicks per user between the variant (M = 0.60, SD = 0.62) and the control group (M = 0.54, SD = 0.37); \textit{t}(447) = 1.24, \textit{p} = .22.

\subsection{Newspaper's Brand Identity}\label{sec:system_impact}
We wanted to make sure we are indeed serving more local content to users. Thus, we analyzed the number of local items served in Taboola's "powered" slots—specific content-serving positions within website sections, such as Sports. Taboola operates three types of sections, powering 3, 10, or 12 slots in each. Table \ref{tab:system_impact} shows the average lift by section type between the local and control variants across a network of more than 100 local news websites.

\begin{table}
  \caption{Average Lift of Local Articles in Taboola-powered sections}
  \label{tab:system_impact}
  \begin{tabular}{cccl}
    \toprule
        Powered Slots&Average Lift\\
    \midrule
        3&43.35\%\\
        10&19.28\%\\
        12&22.23\%\\
    \bottomrule
        Weighted average&\textbf{27.30}\%\\
\end{tabular}
\end{table}

\section{Conclusions and Discussion}

Our offline evaluations demonstrated that LLMs surpass traditional methods in local classification tasks. ChatGPT exhibited exceptional relational knowledge in identifying article locations, even without Knowledge Graph data enrichment. For cases where LLMs aren't utilized, combining NER and Knowledge Graphs remains the most effective approach for detecting implicit locations.

Online A/B tests showed significant increases in local views per user, while overall clicks per user remained unaffected. The lack of significant impact on overall engagement was expected, given the model's bias toward national content due to sparse local data. Production brand identity analysis showed an uplift of 27.3\% in local content served to in-market users, demonstrating that the approach effectively maintains newspapers' brand identity without compromising user engagement.

Knowledge Graph prompt enrichment benefited only 3\% of articles tagged as local, raising questions about the trade-off between the value of KG-LLM integration and its implementation overhead. When examining potential causes of ChatGPT errors, we found errors caused by editor tagging, county-state mismatches, Knowledge Noise, and our oversimplified classification criteria.

We did not run into LLM outdatedness in the current dataset. Nevertheless, this can happen with athlete team changes or mentions of unfamiliar public figures. Since locations rarely change, outdatedness posed only a minor concern. We expect this issue to diminish further as future LLMs receive more frequent training updates or gain direct internet access.

When examining Geographical Ambiguity, we showed that while knowledge graph enrichment can create additional toponym ambiguities through knowledge noise, it can also aid in resolving existing ambiguities, improving classification accuracy.

Regarding hallucinations, we filtered out hallucinated location entities from the data before processing by validating that the locations exist in the world and that cities are located within their suggested states/counties.

To enhance explainability, we implemented the CoT technique, which structures instructions into sequential steps and improves ChatGPT's classification accuracy. While analyzing ChatGPT's explanations could have provided deeper insight into its decisions, this proved impractical in high-scale systems processing 500,000 recommendation requests per second. However, a brief indication of whether Knowledge Graph information contributed positively or negatively to the decision could provide valuable reasoning insights with minimal computational overhead—a possibility worthy of further study, potentially combined with LLM-as-a-Judge paradigms to enhance reasoning capabilities \cite{gu2024survey}.

In order to maintain efficiency, we explicitly instructed ChatGPT to omit reasoning details. Still, enriching prompts with Knowledge Graph data revealed indirect insights into LLM decision-making. These findings warrant further testing with larger samples and could benefit from additional Knowledge Graph information beyond person entities, sports teams, and organizations. As modern LLMs increasingly reveal their reasoning processes (such as in DeepSeek R1 "think out loud" technique \cite{deepseekai2025deepseekr1incentivizingreasoningcapability}), future research should explore how Chain-of-Thought and Knowledge Graph enrichment can further improve reasoning transparency.

Finally, our production-scale pipeline extends beyond local classification. Taboola now extracts diverse insights from articles, enriches features using LLMs and in-house auxiliary models in near real-time, and integrates this data into our recommender system at scale. This enables more personalized user experiences while preserving each newspaper's brand identity.

\begin{acks}
We would like to thank Taboola for enabling to conduct this research, and to Maoz Cohen, Elad Gov-Ari and Gil Chamiel for reviewing it.
\end{acks}

\bibliographystyle{ACM-Reference-Format}
\bibliography{references}

\appendix
\section{Appendix} \label{appendix:appendix_a}
\subsection{ChatGPT Prompt Prefix} \label{appendix:prompt}
\begin{lstlisting}[language=json,firstnumber=1]
You are a newspaper editor. For any given news article's title and description, your task is to identify the associated city, state and country, even when not explicitly stated.

Follow these steps:
First, identify the entities in the title (Person, Sports team, Business, or Organization).
Then, search your training data for the identified entity.
For person names: Determine who they are and extract their associated city, state, and country. For political figures, identify their party's base location.
For sports teams: Find their home city and venue.
For businesses or organizations: Locate their headquarters or main operating location.
If there is a tip from the Knowledge Graph in the form of <entity,relation,location>, use it. 
No abbreviations are allowed, write out the full name of the city, state and country.
No external resources are needed, only your training data.

Additional instructions: 
1. Return output in valid JSON format:
{"title": "{{title}}",
"item_city": "{{city}}",
"item_state": "{{state}}",
"item_country": "{{country}}"}
2. Use "null" for any categories where information is insufficient or unknown.
3. Omit explanations after the output.
4. Include the provided title in your JSON response.
5. Return only the JSON output with no additional text.
6. For multiple locations in a category, select the first one and return a single value for item_city, item_state, item_country.
 Example 1:
    {"title": "Dolphin's Grant DuBose stretchered off, remains hospitalized",
    "description: "The player remains hospitalized Monday for continued observation after suffering a serious head injury."}
    output:
    {"title": "Dolphin's Grant DuBose stretchered off, remains hospitalized",
     "item_city": "miami",
     "item_state": "florida",
     "item_country": "united states",
    }

Example 2:
 {"title": "Governor Ron DeSantis Announces the Focus on Fiscal Responsibility 2025-2026 Budget",
    "description: "Today, Governor Ron DeSantis announced his budget proposal for Fiscal Year (FY) 2025-2026."}
    output:
    {"title": "Governor Ron DeSantis Announces the Focus on Fiscal Responsibility 2025-2026 Budget",
     "item_city": "miami",
     "item_state": "florida",
     "item_country": "united States",
    }
\end{lstlisting}

\subsection{ChatGPT Confusion Matrices} \label{appendix:confusion_matrix}
Table \ref{tab:confusion_matrix_chatgpt} presents the confusion matrices received in both Standalone-ChatGPT and KG-Enriched-ChatGPT.
\begin{table}
\caption{ChatGPT Confusion Matrix}
\label{tab:confusion_matrix_chatgpt}
\begin{tabular}{cccl}
\toprule
    ChatGPT&Label&True Local&True Not Local\\
\midrule
    Standalone&Predicted-Local&1829&415\\
    Standalone&Predicted-Not Local&498&31138\\
    Enriched&Predicted-Local&1897&413\\
    Enriched&Predicted-Not Local&430&31140\\
\bottomrule
\end{tabular}
\end{table}

%%% -*-BibTeX-*-
%%% Do NOT edit. File created by BibTeX with style
%%% ACM-Reference-Format-Journals [18-Jan-2012].

\end{document}